\documentclass{article}
\usepackage{ijcai22}

\usepackage{times}
\usepackage{soul}
\usepackage{url}
\usepackage[hidelinks]{hyperref}
\usepackage[utf8]{inputenc}
\usepackage[small]{caption}
\usepackage{graphicx}
\usepackage{amsmath}
\usepackage{amsfonts}
\usepackage{amssymb}
\usepackage{amsthm}
\usepackage{booktabs}
\usepackage{algorithm}
\usepackage{algorithmic}
\usepackage{color}
\usepackage{multirow}

\title{Planning-inspired Hierarchical Trajectory Prediction for Autonomous Driving}

\author{ Ding Li$^{1, 2}$
\and
Qichao Zhang$^{2, 1}$\and
Zhongpu Xia$^{3}$\and
Kuan Zhang$^{3}$\and
Menglong Yi$^{3}$\and
Wenda Jin$^{3}$\And
Dongbin Zhao$^{2, 1}$
\affiliations
$^1$School of Artificial Intelligence, University of Chinese Academy of Sciences, China\\
$^2$State Key Laboratory of Management and Control for Complex Systems, Institute of Automation, Chinese Academy of Sciences, China\\
$^3$Baidu Inc., China
\emails
\{liding2020, zhangqichao2014, dongbin.zhao\}@ia.ac.cn,\\
\{xiazhongpu01, zhangkuan01, yimenglong, jinwenda\}@baidu.com
}

\begin{document}

\maketitle

\begin{abstract}
 Recently, anchor-based trajectory prediction methods have shown promising performance, which directly selects a final set of anchors as future intents in the spatio-temporal coupled space. However, such methods typically neglect a deeper semantic interpretation of path intents and suffer from inferior performance under the imperfect High-Definition (HD) map. To address this challenge, we propose a novel Planning-inspired Hierarchical (PiH) trajectory prediction framework that selects path and speed intents through a hierarchical lateral and longitudinal decomposition. Especially, a hybrid lateral predictor is presented to select a set of fixed-distance lateral paths from map-based road-following and cluster-based free-move path candidates. {Then, the subsequent longitudinal predictor selects plausible goals sampled from a set of lateral paths as speed intents.} Finally, a trajectory decoder is given to generate future trajectories conditioned on a categorical distribution over lateral-longitudinal intents. Experiments demonstrate that PiH achieves competitive and more balanced results against state-of-the-art methods on the Argoverse motion forecasting benchmark and has the strongest robustness under the imperfect HD map.
\end{abstract}

\section{Introduction}
\begin{figure*}[htbp]
	\centering
		\scriptsize
		\includegraphics*[width=6.1in]{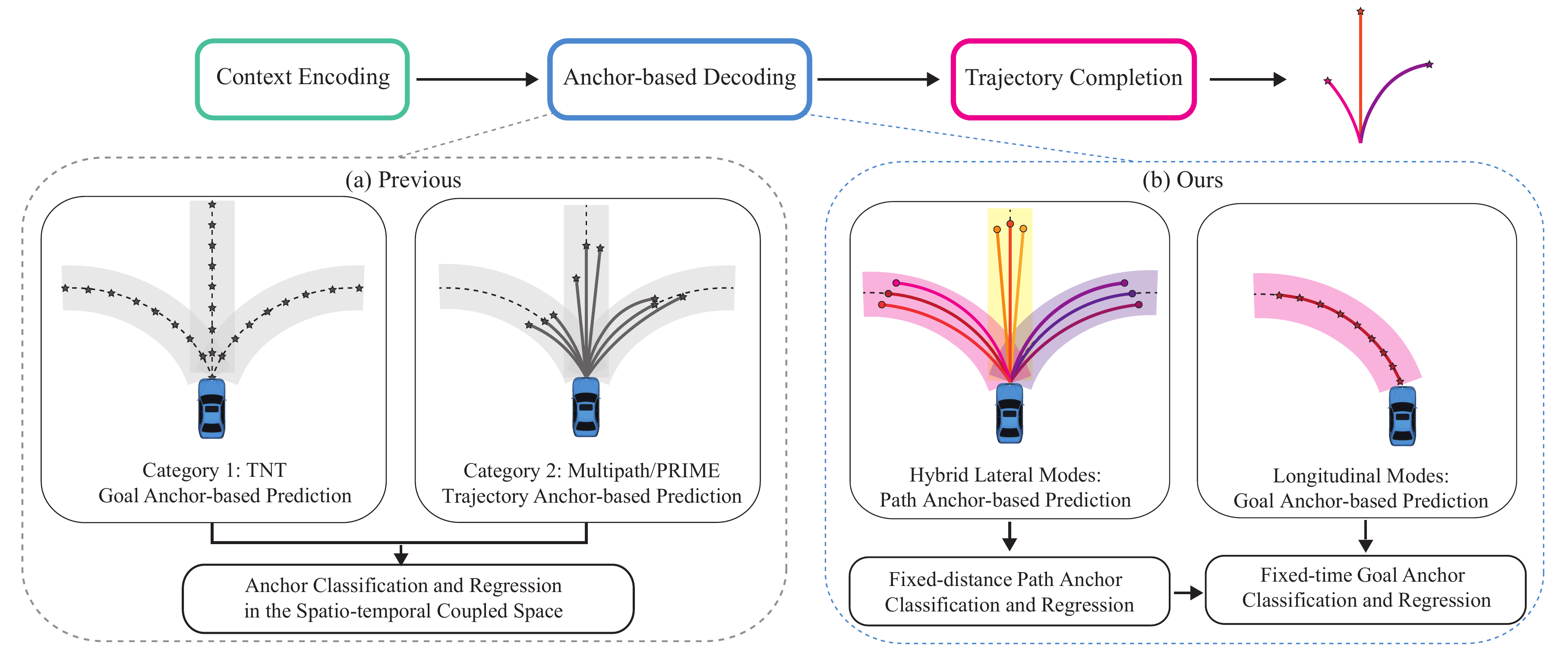}
		\caption{{Comparison of (a) anchor-based prediction methods and (b) our proposed PiH.} Different from existing anchor-based prediction methods (lower left), which select anchors directly in the spatio-temporal space, our PiH (lower right) obtains anchors through a hierarchical lateral and longitudinal decomposition. The lateral path anchors ask the "\textit{where to go}" question and the longitudinal goal anchors  ask the "\textit{when will it arrives}" question.}
		\label{fig:motivation}
   \vspace{-1.3em} 
\end{figure*}
Trajectory prediction is a mission-critical component for autonomous driving, which bridges the upstream perception and downstream planning {\cite{trajGen}}. Based on the perceptual outcome and HD map, it is responsible for inferring future multimodal behaviors of on-road agents (e.g., surrounding vehicles and pedestrians) to facilitate the planning of self-driving vehicles (SDVs) for safe and comfortable social interactions. The challenges of trajectory prediction lie in addressing heterogeneous static and dynamic scene inputs, modeling multi-agent interaction, and reasoning the multimodal output under complex driving scenarios, which can not be handled well based on traditional methods with handcrafted maneuvers \cite{AdamHouenou2013VehicleTP,StephanieLefevre2014ASO} due to the randomness and subjectivity of human behaviors.

The learning-based approaches \cite{WenyuanZeng2019EndToEndIN,VectorNet2020,DenseTNT2021} significantly improve the prediction performance and dominate the popular motion forecasting benchmarks \cite{Argoverse2019,WaymoDataset2021} by leveraging deep neural networks for reasonable scene understanding. To address the high degree of uncertainty, {tremendous efforts are spent on modeling the multimodal output.} One class of approaches predicts multiple future behaviors by sampling from the implicit distribution represented by the latent variables, such as CVAEs \cite{Desire2017}, GANs \cite{AgrimGupta2018SocialGS}, and single-step policy rollout methods \cite{Precog2019}. However, such sampling-based methods hardly output the likelihood of predicted trajectories and expose poor interpretation of latent variables.


Recently, anchor-based methods have demonstrated their superiority in the multimodal motion prediction task.
This class of methods first discretizes the output space with explicit possible anchors, such as handcrafted motion maneuvers \cite{SergioCasas2018IntentNetLT}, future trajectory candidates \cite{MultiPath2019,CoverNet2019,PRIME2021}, goal candidates \cite{TNT2020,WenyuanZeng2021LaneRCNNDR}, and so on. Then, plausible future intentions are obtained by classifying and regressing those prior anchors, and the multimodal trajectories can be achieved conditioned on those intentions. Despite their competitive performance, these anchor-based methods have two main limitations. First, those plausible anchors are directly selected by anchor classification and regression in the spatio-temporal coupled space. It is  challenging to achieve high-quality anchors due to the inherent stochasticity and subjectivity of human behaviors. Second, the phenomenon of HD map missing and mismatching occurs commonly in road testing; thus, these methods may ruin the prediction robustness under the imperfect HD map, i.e., {PRIME \cite{PRIME2021} model-based trajectory anchors and TNT \cite{TNT2020} pre-defined goal anchors may be lost for missing lanes in the HD map.}


To address these limitations, we propose PiH, a hierarchical anchor-based trajectory prediction method to model the output multimodalities, as shown in Figure \ref{fig:motivation}. The contributions of our work can be summarized as follows: {1) Inspired by path-speed decoupled planning \cite{appllo-planning}, PiH is the first work so far to learn a \textit{lateral-longitudinal} decoupled hierarchical manner for anchor-based trajectory prediction, including the \textit{fixed-distance path anchors} (lateral modes) and subsequent \textit{fixed-time goal anchors} (longitudinal modes). 2) A hybrid lateral predictor is designed to select a set of lateral paths from \textit{map-based road-following} and \textit{cluster-based free-move} path candidates, which can improve the prediction robustness under the imperfect HD map. 3) Following Multipath++ \cite{MultiPath++}, we present a simple yet efficient temporal multi-context gating (MCG) encoder to capture time-series relations of historical dynamics. 4) We outperform state-of-the-art methods on the Argoverse motion forecasting benchmark and achieve superior robustness with only a 0.2 $\sim$ 0.3 times increase by comparison with PRIME and DenseTNT that degrades seriously with several times increase under the imperfect map. 

\section{Related Work}

\subsection{Scene Encoding}
To fuse static (e.g., road geometry, lane connectivity, static obstacles) and dynamic (e.g., time-varying traffic lights, dynamic obstacles) inputs, the rasterized representation is used for scene context encoding with convolutional neural networks (ConvNets) \cite{MultiPath2019,WenyuanZeng2019EndToEndIN},  
which requires manual specifications such as the color-coded attributes. VectorNet \cite{VectorNet2020} provides a vectorized representation for the HD map and multi-agent dynamics with graph neural networks (GNNs), which avoids lossy rendering and computationally intensive ConvNet encoding steps. 
Based on the vectorized representation, MultiPath++ \cite{MultiPath++} designs a MCG fusion component for efficient scene encoding. 
Recently, Transformer {\cite{Transformer2017,STPrediction}} has become a
popular choice for interaction-aware motion modeling based on the attention mechanism. SceneTransformer \cite{SceneTrans2021} is
designed for the multi-agent prediction task with the Transformer-based scene encoder, which handles the interactive modeling among timesteps,
agents, and road graph elements in a unified way.

\subsection{Anchor-based Multimodal Trajectory}
\begin{figure*}[htbp]
	\centering
	\begin{center}
		\scriptsize
		\includegraphics*[width=7.1in]{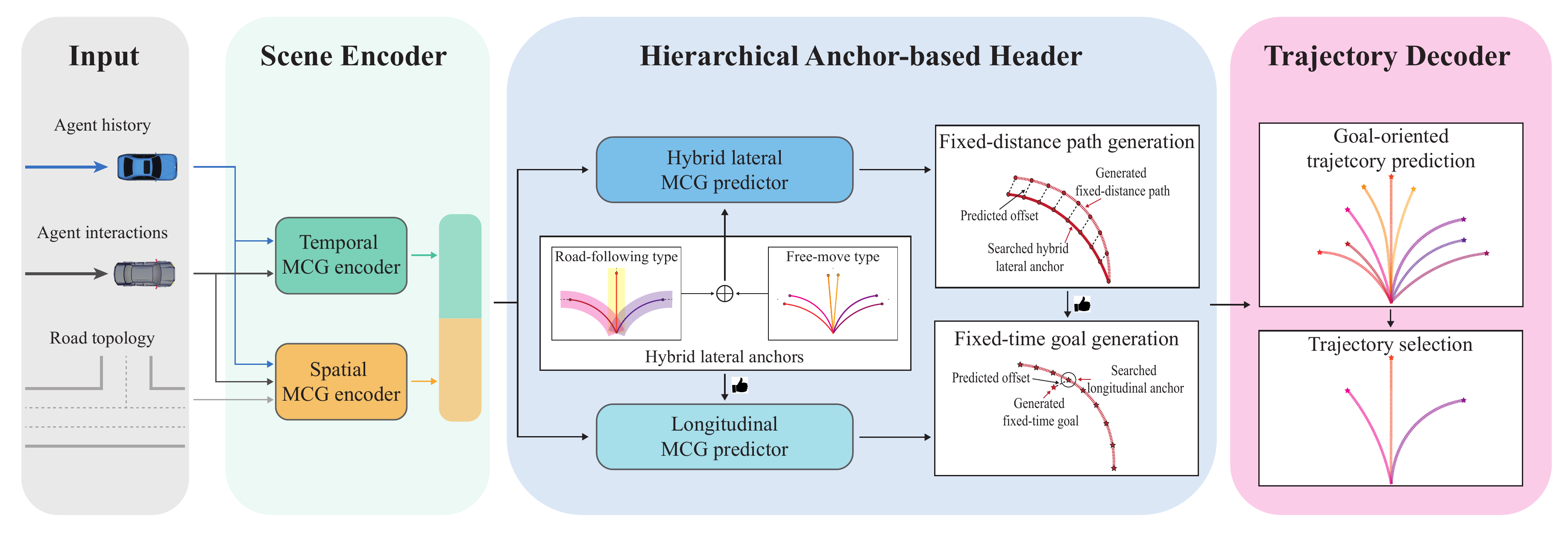}
		\caption{An overview of PiH. We utilize the MCG-based scene encoder as the feature extraction module to model the temporal and spatial relationship among agents and map information. Then, a hierarchical anchor-based header is used to classify and regress the potential lateral fixed-distance paths and longitudinal fixed-time goal candidates. Finally, with the goal position, we adopt a trajectory decoder to predict and subsequently select multimodal goal-oriented trajectories. }
		\label{fig:framework}
	\end{center}
   \vspace{-1.3em} 
\end{figure*}

Existing works design different types of anchors that discretize the multimodal output space over the fixed-time prediction horizon and achieve promising performance. 
For example, MultiPath \cite{MultiPath2019} and CoverNet \cite{CoverNet2019} generate pre-defined, static trajectory candidates as hypothesis anchors. Further,
TNT \cite{TNT2020} generates goal-oriented trajectories to diversify the prediction modes, with the goal anchor candidates sampled from the HD map. Besides, since the topological structure of lanes can be considered as guidance for the motion of drivers, a vast majority of recent works leverage a set of instance-level lane entities as spatial anchors to generate plausible multimodal trajectories \cite{LaPred2021}. Different from these anchor types above, \cite{mmTransformer2021} constructs a region-based training method to cover all the possible prediction modes with limited training samples, where the surrounding space is divided into a small number of regions.  

\subsection{Prediction and Planning}
Since prediction and planning are closely intertwined, some existing works attempt to incorporate planning ideas into the prediction or integrate planning and prediction.  
To improve the kinematic feasibility of predicted trajectories, DKM \cite{HenggangCui2019DeepKM} embeds a kinematic vehicle motion model in the output layer of the backbone prediction network. To ensure kinematic and environmental feasibility, TPNet \cite{TPNet2020} first predicts a rough future endpoint, then generates possible future trajectory proposals conditioned on the predicted endpoint using the polynomial curve. PRIME \cite{PRIME2021} directly searches a set of reachable paths and generates possible future trajectory anchors in a planning manner. Besides, the novel planning-prediction-coupled frameworks are introduced to make predictions conditioned on ego intentions \cite{Precog2019} or complete planned trajectories \cite{HaoranSong2020PiPPT,Trajectron++2020}, which is capable of providing accurate interaction-aware trajectory prediction.

\subsection{Different from Previous Works}
Our work fully utilizes superiorities from vectorized-based \cite{MultiPath++}, trajectory/goal-based \cite{MultiPath2019,TNT2020}, and planning-based \cite{PRIME2021} methods. Besides, our work has three appealing benefits that previous works disregard: 1) Compared with trajectory/goal-based methods directly performing intent estimation in a spatio-temporal coupled space, PiH enables the model to integrate future intents from both lateral and longitudinal aspects in a hierarchical decoupled manner. In addition to reducing the difficulty of the anchor selection, a more faithful interpretation of path intents can be obtained to facilitate the downstream planning. 2) Compared with those only taking map-based or cluster-based priors into account, our proposed lateral predictor, with hybrid manners of road-following and free-move, is more generalizable and robust in different scenarios, especially under the imperfect HD map. 3) Different from the multi-stage planning-based method PRIME, PiH can be learned in an end-to-end manner.


\section{Problem Formulation}
\subsection{Input Multimodality}
Scene context $\boldsymbol{X}$ is composed of a highly heterogeneous mixture of static and dynamic inputs, including agent history $\boldsymbol{X}_{tar}$, agent interactions $\boldsymbol{X}_{obs}$, and road topology $\boldsymbol{X}_{road}$. For simplicity, the following symbols are defined: $T$ denotes the number of past time steps considered in the historical trajectory, and $D$ denotes the number of waypoints in the road segment. Following VectorNet \cite{VectorNet2020}, we use polylines to represent the historical trajectory and road segment, with $P_{obs}$ and $P_{road}$ denoting the number of the above around the target vehicle. Additionally, $I_{tar}$, $I_{obs}$, and $I_{road}$ denote the feature dimensions of the target vehicle, neighboring agents, and road segments, respectively. PiH makes predictions based on the following input modalities:
\begin{itemize}
    \item \textit{Agent history} $\boldsymbol{X}_{tar}$ contains a sequence of past states for the target vehicle $[1, T, I_{tar}]$. Towards each time step $t$ $\in$ $T$, the state features include position, velocity statistics, and acceleration.
    \item \textit{Agent interactions} $\boldsymbol{X}_{obs}$ can be represented by the interaction tensor $[P_{obs}, T, I_{obs}]$. For each neighboring agent $p_{obs}$ $\in$ $P_{obs}$, the state features are extracted in the same format as the \textit{Agent history} above, but transformed into the reference frame of our target vehicle.
    \item \textit{Road topology} $\boldsymbol{X}_{road}$ can be summarized by the roadgraph $[P_{road}, D, I_{road}]$, containing $P_{road}$ road segments closest to the target vehicle. We further approximate each road segment $p_{road}$ $\in$ $P_{road}$ as a sequence of waypoints, where the waypoint features include the position and some attributes of road segments, such as traffic control, turn direction, and so on.
\end{itemize}

\subsection{Hierarchical Output Multimodality}
To account for the multimodality of the distribution of future trajectories, we output a set of $K$ trajectories $\{{\boldsymbol{Y}}^{k}\}_{k \in [1, K]}$ for the target vehicle over a fixed prediction horizon of time steps. This multimodality fuses two dimensions, including lateral paths (“\textit{where to go}”) and longitudinal speeds (“\textit{when will it arrives}”). Different from directly selecting intent anchors in the spatio-temporal space, we decompose this task into classifying and regressing \textit{fixed-distance lateral path anchors} and \textit{fixed-time longitudinal goal anchors} hierarchically. 
{The lateral path generation stage alleviates the uncertainty from randomness and subjectivity along the time axis, which decreases the difficulty of the intent classification task.} Furthermore, the generated set of lateral paths efficiently reduces the searching space for the longitudinal intention estimation.

Specifically, the hierarchical prediction strategy predicts $L$ fixed-distance paths with probabilities and then predicts $M$ fixed-time goals as longitudinal modes for each lateral mode. As a result, the number of lateral-longitudinal modes is $N = L \cdot M$, and the probability of each one is the product of the lateral mode probability and the longitudinal mode probability conditioned on its underlying lateral fixed-distance path:
\begin{equation}
    p\big(\boldsymbol{Y} | \boldsymbol{X}\big) =  \sum_{i=1}^L \sum_{j=1}^M \underbrace{p\big(\boldsymbol{\mathcal{P}}_i | \boldsymbol{X}\big)}_{lateral} \underbrace{p\big(\boldsymbol{g}_i^j | \boldsymbol{\mathcal{P}}_i, \boldsymbol{X}\big)}_{longitudinal} p\big(\boldsymbol{Y} | \boldsymbol{g}_i^j, \boldsymbol{\mathcal{P}}_i, \boldsymbol{X}\big),
\end{equation}%
where $\boldsymbol{g}_i^j$ denotes the \textit{j}-th longitudinal goal candidate from the \textit{i}-th lateral fixed-distance path $\boldsymbol{\mathcal{P}}_i$.  Further, the $K$ trajectories can be obtained from $N$ predictions via the trajectory selection. $\mathcal{P}_1$


\section{Method}
Figure \ref{fig:framework} depicts a high-level overview of the proposed PiH model. Given multimodal inputs in a scene, PiH seeks to 1) fuse the scene context with the temporal and spatial MCG encoders; 2) provide a compact weighted set of fixed-distance paths and fixed-time goals based on the hierarchical anchor-based header; 3) predict and select the final trajectories conditioned on a categorical distribution over lateral-longitudinal intents through the trajectory decoder.
\subsection{Scene Encoder}
This module focuses on interactive representation under temporal and spatial dimensions. It consists of two MCG-based encoders to learn past trajectories and scene information, respectively. The structure details can be found in Appendix.

\subsubsection{Spatial MCG Encoder} 
We first independently encode agent history $\boldsymbol{X}_{tar}$, agent interactions $\boldsymbol{X}_{obs}$, and road topology $\boldsymbol{X}_{road}$ via 1D conv layers, and project the spatial features from the initial dimension to $[1, T, H]$, $[P_{obs}, T, H]$, and $[P_{road}, D, H]$, respectively, where $H$ is the embed dimension of scene encoder. Then, the max-pooling layer is adopted across 2\textit{rd}-dimension to get the aggregated spatial features $\boldsymbol{C}_{tar}$, $\boldsymbol{C}_{obs}$, and $\boldsymbol{C}_{road}$, with the size of $[1, H]$, $[P_{obs}, H]$, and $[P_{road}, H]$, respectively. Finally, we fuse agent interaction embedding $\boldsymbol{C}_{obs}$ and road topology embedding $\boldsymbol{C}_{road}$ with stacked MCG blocks as follows:
\begin{equation}
\begin{split}
\boldsymbol{C}'_{tar}, \boldsymbol{C}'_{obs} &= {\rm{MCG}}_N({\rm context}=\boldsymbol{C}_{tar}, {\rm state}=\boldsymbol{C}_{obs}), \\
\boldsymbol{Z}_{spat}, \boldsymbol{C}'_{road} &= {\rm{MCG}}_N({\rm context}=\boldsymbol{C}'_{tar}, {\rm state}=\boldsymbol{C}_{road}).
\end{split}
\end{equation}%
Thus, we consider $\boldsymbol{Z}_{spat}$ as the final spatial feature.

\subsubsection{Temporal MCG Encoder} 
Different from the spatial MCG encoder, this module especially extracts temporal relations between multi-agents. The temporal features are purely encoded from agent history $\boldsymbol{X}_{tar}$ and agent interactions $\boldsymbol{X}_{obs}$ using the Deep $\&$ Cross Network \cite{DCN-v2}. During concatenation and max-pooling across 1\textit{st}-dimension, we obtain the compact temporal feature $\boldsymbol{S}_{tar}$ with the size of $[T, H]$. The stacked MCG blocks are employed to fuse the history embedding into the target vehicle:
\begin{equation}
\boldsymbol{Z}_{temp}, \boldsymbol{S}'_{tar} = {\rm{MCG}}_N({\rm context}=\boldsymbol{C}_{tar}, {\rm state}=\boldsymbol{S}_{tar}).
\end{equation}%
Here, we consider $\boldsymbol{Z}_{temp}$ as the final temporal feature.

Both $\boldsymbol{Z}_{spat}$ and $\boldsymbol{Z}_{temp}$ are concatenated and fed into the Multi-Layer Perception (MLP) layer, outputting the final embedding $\boldsymbol{Z}_{tar} = {\rm MLP}([\boldsymbol{Z}_{spat}; \boldsymbol{Z}_{temp}])$ that aggregates spatial-temporal features from the driving scene.

\subsection{Hierarchical Anchor-based Header}
This section aims to seek a hierarchical policy that samples roll-outs of the intents correspond to likely lateral paths and longitudinal goals  for the target vehicle in the future.

\subsubsection{Hybrid Lateral MCG Predictor}

 To improve the robustness of the prediction model under the imperfect HD map,  we integrate two types of lateral modes including \textit{road-following} and \textit{free-move} path candidates denoted by $\boldsymbol{\mathcal{A}}_{\mathcal{P}} = \{\boldsymbol{a}_{\mathcal{P}, i}\}_{i=1}^{L}$. The difference between the two types of fixed-distance path candidates is that road-following paths are the centerlines of potential successor roads based on the HD map, while the free-move paths are pre-clustered based on the trajectory dataset {(see Appendix for details}). Note that each fixed-distance lateral anchor ${\boldsymbol{a}_{\mathcal{P}, i}}$ is made up of $M$ waypoints with 2-dimensional position information.

Inspired by the anchor-based prediction method \cite{MultiPath2019}, 
we model the potential fixed-distance paths via a set of $L$ hybird lateral anchors with continuous offsets: $\boldsymbol{\mathcal{P}} = \{\boldsymbol{\mathcal{P}}_i\} = \{\boldsymbol{a}_{\mathcal{P}, i} + \Delta \boldsymbol{a}_{\mathcal{P}, i}\}_{i=1}^L$. Concretely, we encode lateral anchors $\boldsymbol{\mathcal{A}}_{\mathcal{P}}$ by using the MLP and max-pooling layers from size $[L, M, 2]$ to $[L, H]$. Then the final scene embedding $\boldsymbol{Z}_{tar}$ is used as context in stacked MCG blocks that operate on the set of lateral anchor embedding $\boldsymbol{E}_{\mathcal{P}}$, with a final MLP that predicts a discrete distribution over fixed-distance paths and their most likely offsets:
\begin{equation}
\resizebox{1.0\linewidth}{!}{$
\big(p(\boldsymbol{\mathcal{P}}_i | \boldsymbol{X}), \Delta \boldsymbol{a}_{\mathcal{P}, i}\big)_{i=1}^L = {\rm MLP}({\rm{MCG}}_N({\rm context}=\boldsymbol{Z}_{tar}, {\rm state}=\boldsymbol{E}_{\mathcal{P}})).$}
\end{equation}%
Besides, we normalize the lateral probabilities using a softmax layer to output the lateral policy.

\subsubsection{Longitudinal MCG Predictor}
The target vehicle may exhibit diverse longitudinal speed intents on a lateral path such as slowing down and speeding up, which can be captured by the goal at the final time step into the future. Instead of performing goal prediction in the unrestricted map searching space, we properly design the longitudinal anchor searching space $\boldsymbol{\mathcal{A}}_{g, i}=\{\boldsymbol{a}_{g,i}^j\}_{j=1}^M$ by collecting waypoints along the fixed-distance path $\boldsymbol{\mathcal{P}}_{i}$ provided by the lateral predictor.

Following TNT \cite{TNT2020}, we estimate the probability and the position offset for each longitudinal anchor to produce the goal's location.  We first get the initial feature $\boldsymbol{E}_{g, k}$ of goals with the size of $[M, H]$ by encoding waypoints $\boldsymbol{\mathcal{A}}_{g, k}$ sampled along the most likely lateral path $\mathcal{P}_k$ using MLP. Further, we select the updated lateral state feature $\boldsymbol{E}'_{\mathcal{P}, k}$ as the context of stacked MCG blocks to extract the local information between the goals and the lateral path candidate. Given such representations above, a final MLP is applied to output the goals' distribution and positional offsets:
\begin{equation}
\resizebox{1.0\linewidth}{!}{$
\big(p(\boldsymbol{g}_k^j|\boldsymbol{\mathcal{P}}_k, \boldsymbol{X}), \Delta \boldsymbol{a}_{g, k}^j\big)_{j=1}^M = {\rm MLP}({\rm{MCG}}_N({\rm context}=\boldsymbol{E}'_{\mathcal{P}, k}, {\rm state}=\boldsymbol{E}_{g, k})).$}
\end{equation}%
Similarly, all $M$ probabilities of goals in each lateral path will be normalized via a softmax layer. Note that we apply a teacher forcing technique \cite{teacherforcing} by feeding the ground truth lateral path during training.
\subsection{Trajectory Decoder}
For each combination over the lateral path and longitudinal goal, we append a prediction header for generating future trajectories. Then, an effective trajectory selection algorithm is used to further improve coverage and boost prediction performance with a limited budget on trajectories.
\subsubsection{Goal-oriented Trajectory Predictor}
This step is to complete each trajectory conditioned on the predicted goals that integrate the lateral and longitudinal uncertainties. We first calculate the feature of each goal by aggregating the scene embedding $\boldsymbol{Z}_{tar}$ and the updated lateral embedding $\boldsymbol{E}'_{\mathcal{P}, k}$, and the longitudinal embedding $\boldsymbol{E}'^l_{g, k}$, then pass it to the trajectory prediction header realized by MLP. When training the trajectory predictor, we also use the teacher forcing techniques \cite{teacherforcing} by providing the ground truth goal as the target.

\subsubsection{Trajectory Selector}
At inference time, we first call the hybrid lateral MCG predictor and select $\Bar{L} (\leq L)$ fixed-distance paths with higher probabilities. Second, for each predicted lateral path, we generate $\Bar{M} (\leq M)$ fixed-time goals using the longitudinal MCG predictor. Third, we adopt the goal-oriented trajectory predictor to produce $\Bar{L} * \Bar{M}$ trajectories simultaneously. Fourth, we apply the K-means algorithm \cite{kmeans} to generate $K$ trajectory clusters from the total $\Bar{L} * \Bar{M}$ predicted goal-oriented trajectories. Within each cluster, we average all trajectories in the cluster to output the final trajectory, and use the sum of their probabilities as the score. Finally, all $K$ probabilities of trajectories will be normalized.

\subsection{Learning}
We jointly train the hierarchical model with a loss containing the lateral prediction, longitudinal prediction, and trajectory regression. Specifically, we use 
\begin{align}
\mathcal{L} = \alpha \mathcal{L}_{late} + \beta \mathcal{L}_{long} + \gamma \mathcal{L}_{traj},
\end{align}%
where $\alpha$, $\beta$, and $\gamma$ are hyper-parameters determining the relative weights of different tasks. As both lateral and longitudinal prediction tasks are realized by intent classification and offset regression, we simply adopt a binary cross entropy loss for intent classification and a smooth-L1 loss for offset regression. As well, $\mathcal{L}_{traj}$ is realized by a smooth-L1 loss.
\section{Experiments}
In this section, we introduce the dataset benchmark and the details of our model, followed by a series of experiments to demonstrate the effectiveness of PiH against several state-of-the-art methods.

\subsection{Dataset}
We train and validate PiH on the large-scale Argoverse motion forecasting dataset \cite{Argoverse2019}, a popular benchmark commonly used for single vehicle trajectory prediction. This dataset contains 205942 scenarios in the training set and 39472 sequences in the validation set. Given 2 seconds of observed trajectories, sampled at 10 Hz, as well as the HD map data such as lane centerlines, we need to predict the next 3 seconds of future movements of the target vehicle tagged with 'agent'.

\subsection{Metrics}
We follow the Argoverse benchmark and use minimum average displacement error (minADE), minimum final displacement error (minFDE), and miss rate (MR) to evaluate the qualities of the final $K=6$ predicted trajectories. The metric minADE measures the average $\mathcal{L}_2$ error between the best-predicted trajectory and the ground-truth trajectory over all future time steps, while minFDE measures the $\mathcal{L}_2$ error at the final time step. The best-predicted trajectory is the one that has the minimum endpoint displacement error. MR is the ratio of scenarios where none of the $K=6$ predicted trajectories is within a certain threshold (2m) of ground truth according to the endpoint displacement error.

On the other hand, in order to focus on the lateral pattern and also reduce the impact of randomness and subjectivity of human behaviors, we propose the lateral metrics. Since the final predicted trajectories heavily depend on the lateral paths, we simply compute the displacement error (DE) at equal distances between the top1 predicted trajectory and lateral ground truth, with the distance interval set as 10m. Note that the top1 predicted trajectory is selected by a categorical estimation ranking over lateral-longitudinal intents.

\subsection{Implementation Details}
\subsubsection{Lateral-Longitudinal Candidate Sampling}
Since the training and validation set has a larger proportion of sequences whose future trajectory lengths are within 60m, we set the length of the lateral path to be predicted with a fixed distance of 60m for better generalization. In the hierarchical model, the size of hybrid lateral anchors is set to $L=80$, including $16$ successor roads and $64$ free-move path clusters. Then, we sample $M=30$ longitudinal goal candidates which are densely distributed on these lateral intents, with the sampling density set to 2m.
\subsubsection{Training Details}
As for the multimodal input, we sample $P_{obs}=32$ neighboring obstacles and $P_{road}=128$ road segments with a distance less than 100m from the target vehicle. Both the scene encoder and hierarchical anchor-based header have $N=6$ multi-context gating (MCG) layers. We set all the embed dimensions to $H=64$. The loss weights are $\alpha=8.0$, $\beta=4.0$, $\gamma=0.4$. Our PiH model is trained with a batch size of 16 for 50000 steps on 4 Nvidia RTX 2080 Ti GPUs. We use the Adam optimizer and a cosine annealing learning rate scheduler \cite{warmup}, with the initial learning rate set to 1e-4.

\subsection{Performance}
\subsubsection{Results on Benchmark}
As shown in Table \ref{tab:benchmark}, we provide detailed quantitative results of our PiH model on the Argoverse validation set as well as other public state-of-the-art (SOTA) methods. It shows that our PiH model can achieve competitive and more balanced results over all the evaluation metrics compared with SOTA methods. In contrast with R2P2 \cite{r2p2} and DiversityGAN \cite{diversitygan} that implicitly represent the output modalities as latent variables, PiH achieves significant performance gain on the metrics of minADE and minFDE. Compared to the recent HLSTF with hierarchical latent variables, a better minFDE is obtained for PiH. More importantly, PiH can provide explicit lateral-longitudinal proposals, which are grounded in physical entities that are interpretable.
From another view, DenseTNT \cite{DenseTNT2021} and PRIME \cite{PRIME2021} also adopt explicit proposals, such as goal candidates and trajectory anchors, to predict multimodal trajectories. The proposed method demonstrates its superiority beyond PRIME in minADE and minFDE. Compared to DenseTNT, we can see that the goal-based prediction has a certain advantage in minFDE. {Instead, we observe that PiH can exceed DenseTNT in minADE, benefiting from the deployment of high-level lateral paths. }

\begin{table}[htbp]
\footnotesize
    \centering
    \vspace{0.05cm} 
    \begin{tabular}{l|ccc}
        \toprule
        Methods  & minADE & minFDE & MR \\
        \midrule
        DESIRE \cite{desire}  & 0.92    & 1.77    & 0.18      \\
        R2P2 \cite{r2p2} & 1.40    & 2.35  & -        \\
        Multipath \cite{MultiPath2019} & 0.80 & 1.68 & 0.14 \\
        LaneAttention \cite{laneattention} & 1.05 & 2.06 & -\\
        TNT \cite{TNT2020} & 0.73 & 1.29 & {0.09} \\
        DATF \cite{DATF} & 0.92 & 1.52 & - \\
        DiversityGAN \cite{diversitygan}  & 1.13 & 2.20 & - \\
        PRIME \cite{PRIME2021} & 0.92 & 1.30 & {{0.08}} \\
        LaPred \cite{lapred} & {0.71} & 1.44 & -\\
        HYPER \cite{hyper} & 0.72 & 1.26 & -\\
        LaneRCNN \cite{lanercnn} & 0.77 & {{1.19}} & {\textbf{0.08}} \\
        DenseTNT \cite{DenseTNT2021} & 0.73  & {\textbf{1.05}} & {0.10}\\
        HLSTF \cite{HLSTrajForecast} &{\textbf{0.65}}  & 1.24 & -\\
        PiH (ours) &\textit{\textbf{0.70}} & \textit{\textbf{1.20}} & \textit{\textbf{0.11}} \\
        \bottomrule
    \end{tabular}
    \caption{Comparison of SOTA methods on Argoverse validation set. }
    \label{tab:benchmark}
\end{table}

\noindent\textbf{Lateral Performance}

\noindent Further, we assess the necessity of introducing an extra lateral predictor for our model by comparison with PRIME and DenseTNT in the aspect of lateral metrics. In Table \ref{tab: Argoverse lateral metrics}, we see that PiH has the superior performance on all lateral metrics, even for the one without free-move modes. Despite the prediction errors of final goal-oriented trajectories, our model outperforms DenseTNT that disregards the lateral prediction, which explains the PiH’s advantage in reducing minADE metric by the  hierarchical lateral and longitudinal decomposition. More notably, PiH achieves a significant gain (12 $\sim$ 13\%) in displacement error at the longer distance horizon (50m\_DE). Furthermore, the gap between PiH and the comparison models sharply widens as the horizon increases, which indicates its benefits for long-term lateral prediction and provides better generalization for the subsequent longitudinal prediction.
\begin{table}
\footnotesize
    \centering
    \vspace{0.05cm} 
    \begin{tabular}{c|ccccc}
        \toprule
    \multicolumn{1}{c|}{\multirow{2}{*}{Methods}} & \multicolumn{5}{c}{DE}  \\ \cline{2-6}
    \multicolumn{1}{c|}{}                         & \multicolumn{1}{c}{10m} & \multicolumn{1}{c}{20m} & \multicolumn{1}{c}{30m} & \multicolumn{1}{c}{40m} & \multicolumn{1}{c}{50m} \\
        \midrule
        DenseTNT  & 0.77 & 0.91 &1.00 & 1.08 &1.25\\
        PRIME & 0.81 &0.91 & 0.99 & 1.06 &1.21\\
        \midrule
        PiH w/o free-move & 0.77 &0.88 & \textbf{0.94} & 0.99&1.25\\
        PiH w/ free-move & \textbf{0.75} &\textbf{0.87}  & 0.95 & \textbf{0.98}& \textbf{1.06}\\
        \midrule
        Samples Number & 35121 &24821 & 14799 & 6113 &1038 \\
        \bottomrule
    \end{tabular}
    \caption{Comparison of lateral performance.}
    \label{tab: Argoverse lateral metrics}
\end{table}
\subsection{Ablation Studies}
\subsubsection{Impact of Each Component}
\noindent We conduct an ablation study for our PiH on the Argoverse validation set to evaluate and analyze the contributions of our proposed components to the final performance. We take the goal anchor-based trajectory method as the baseline by utilizing the spatial MCG encoding module as the backbone. And then, we add the additional components to the baseline model gradually. As shown in Table \ref{tab:Ablation}, each component can improve the prediction accuracy to a certain degree. First, the temporal MCG encoding module plays a crucial role in trajectory predictions since inferring the future motions in highly dynamic traffic scenarios heavily relies on the time-series historical information. Second, we note that introducing lateral estimation has a significant impact on performance improvements, which effectively eases the effect of randomness and subjectivity during the prediction of lateral-level path prediction. Further, the production of lateral paths relieves the computation burden of selecting plausible longitudinal-level goal candidates over time. Lastly, the results also show that consideration of free-move lateral intents is more crucial than purely using potential successor roads as lateral anchors in improving the final prediction accuracy.

\begin{table}[htbp]
  \footnotesize
  \centering
    \begin{tabular}{c|cc|ccc}
    \toprule
    \multirow{2}{*}{\scriptsize{Temporal MCG}} & \multicolumn{2}{c|}{\scriptsize{Lateral Estimation}}  & \multirow{2}{*}{minADE} & \multirow{2}{*}{minFDE} & \multirow{2}{*}{MR}  \\
    & \tiny{Road-follow} & \tiny{Free-move} & & &     \\
    \midrule
       & &  &0.78    &1.39    &0.14    \\
     \checkmark & & &0.75      &1.35     &0.14   \\
     \checkmark &\checkmark & &0.73  &1.29  & 0.13 \\
     \checkmark  &\checkmark &\checkmark  & \textbf{0.70} & \textbf{1.20} & \textbf{0.11} \\
    \bottomrule
    \end{tabular}%
    \caption{Ablation studies on the components of our PiH model.}
  \label{tab:Ablation}%
\end{table}%



\begin{figure*}[t]
	\centering
		\scriptsize
		\includegraphics*[width=5.3in]{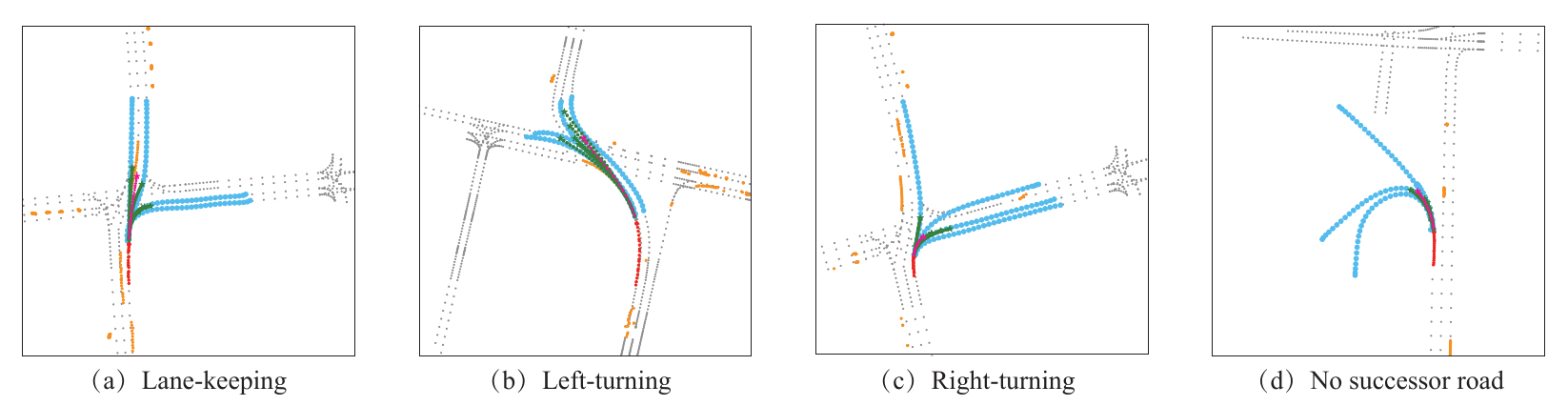}
		\caption{Qualitative results of PiH in 4 typical highly challenging traffic scenarios, such as (a) lane-keeping, (b) left-turning, (c) right-turning, and (d) no successor road. For each scenario, PiH estimates the high-scoring lateral fixed-distance paths in blue, built upon which we predict $K=$6 longitudinal goal-oriented trajectories in green. The ground truth future trajectory is shown in deep pink.
  }
		\label{fig:results}
   \vspace{-1.3em} 
\end{figure*}

\subsubsection{Impact of Lateral Anchor Density}
\noindent One of the key hyper-parameters of our PiH model is the choice of the number of free-move lateral anchors, as it is always the case with methods that rely on K-means clustering. We train several PiH models with different numbers of free-move lateral anchors and evaluate the prediction performance on the validation set, as shown in Table \ref{tab:Lateral Anchor Density}. It indicates that a higher lateral anchor density leads to better performance before the saturating point.

\begin{table}[htbp]
  \footnotesize
  \centering
    \begin{tabular}{cc|ccc}
    \toprule
    \multicolumn{2}{c|}{\scriptsize{Lateral Anchor Density}}  & \multirow{2}{*}{minADE} & \multirow{2}{*}{minFDE} & \multirow{2}{*}{MR}  \\
    \scriptsize{Road-following} & \scriptsize{Free-move} & & &     \\
    \midrule
     16 &28 &0.73  &1.29  &0.13   \\
     16 &38 &0.73  &1.27  &0.13 \\
     16 &48 &0.72 &1.25  &0.12 \\
     16 &64 &\textbf{0.70} &\textbf{1.20} & \textbf{0.11}\\
    \bottomrule
    \end{tabular}%
    \caption{Performance comparison on lateral anchor densities.}
  \label{tab:Lateral Anchor Density}%
\end{table}%

\subsection{Robustness under Imperfect Maps}

\noindent When encountering some challenging situations where the map details are lost or mismatched in some time steps, we require the prediction model to robustly handle the imperfect map instead of being restricted to limited map inputs. To measure the robustness degree of the models (ours, DenseTNT, and PRIME), we randomly drop some map locations around the ground truth in the validation set and then make inferences using original network structures. The map drop rate is set from 0 - 10\% for each data sequence in validation. Intuitively, we plot minADE and minFDE as a function of the drop rate horizon in Figure \ref{fig:imperfect robustnesss results}. Since these comparison models commonly share the HD map to build future intents, the prediction performance exhibits downward trends to a certain degree at the initial drop rate setting. Along with the drop rate increases, we surprisingly notice that our PiH model performs stably with only a 0.2 $\sim$ 0.3 times relative increase on these metrics, while the other methods degrade seriously with several times  relative increase.
This should be attributed to our designed hybrid lateral prediction module, which considers  cluster-based free-move lateral modes to promote scenario adaptability and prediction robustness. Besides, compared to the cluster-based trajectory anchor-based method Multipath \cite{MultiPath2019} shown in Table \ref{tab:benchmark}, PiH still achieves better performance even with 10\% map details lost. 

\subsection{Interpretable Gains}
\noindent In Figure \ref{fig:results}, we present some qualitative results of PiH on the Argoverse validation set. On one hand, we can see that PiH follows the hierarchical prediction strategy and demonstrates accurate, reasonable, and good multimodalities over lateral and longitudinal future intents.
From (a) to (c), when it's approaching an intersection, our model naturally captures both the lane-keeping and turning lateral modes with different velocities (longitudinal modes).
Moreover, as shown in (d), when there is no successor road, we predict future behaviors naturally matching the free-move lateral modes, showing the effectiveness of our hybrid lateral prediction module.

\begin{figure}[htbp]
	\centering
		\scriptsize
		\includegraphics*[width=3in]{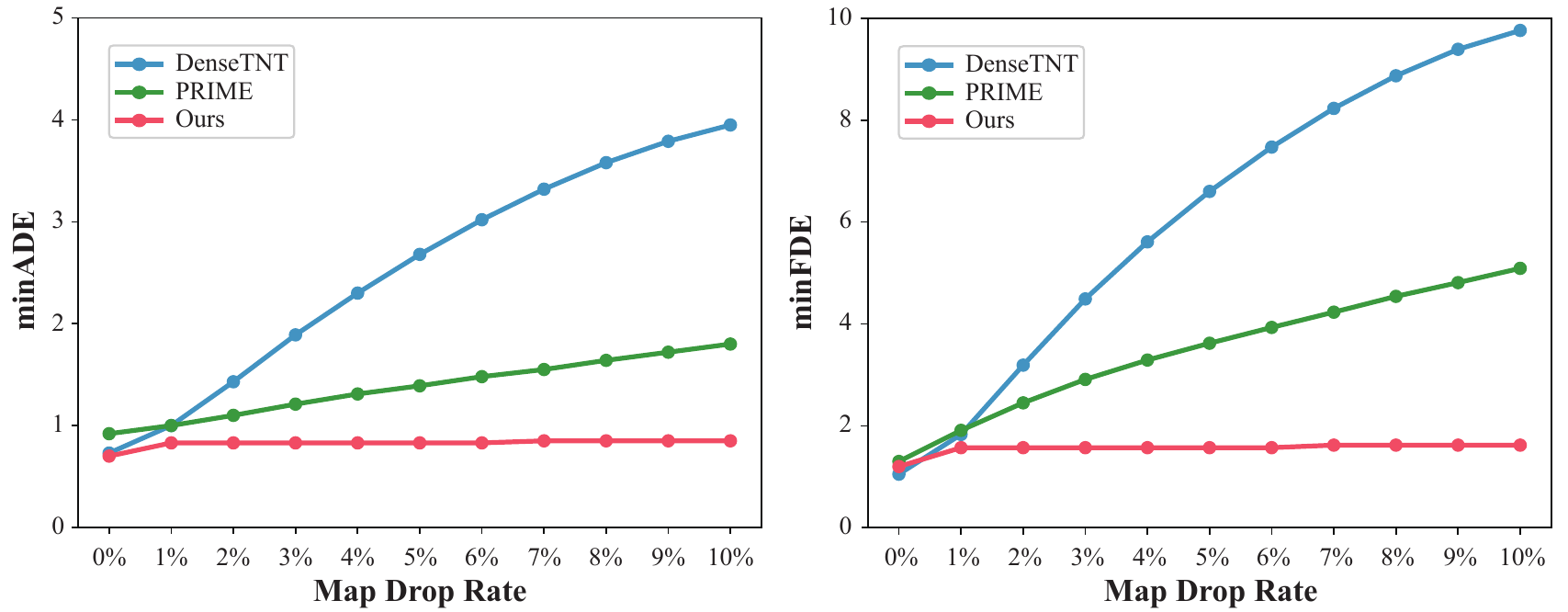}
		\caption{Comparison of prediction robustness under imperfect map. }
		\label{fig:imperfect robustnesss results}
   \vspace{-1.3em} 
\end{figure}

On the other hand, the decomposability of our PiH offers a more fundamental interpretation of the lateral path to facilitate downstream planning. Compared with trajectory-based intentions, lateral-based behaviors equips with better tracking ability, keeping smooth and consistent in multiple successive temporal frames. Based on such advantages, the planning module can successfully identify the underlying interactive agents around the SDVs by detecting overlapping lateral intentions and further making the interactive planning.

\section{Conclusion}
In this paper, we propose PiH, a novel framework for multimodal motion prediction, which hierarchically models the future intents by a factorization of lateral fixed-distance paths and longitudinal fixed-time goal candidates. Built upon this framework, we propose a hybrid lateral prediction module to cover the road-following and free-move path modes. Furthermore, this dedicated lateral design promotes efficient prediction for the subsequent longitudinal estimation and relieves high dependence on the HD map.
Experiments demonstrate that our PiH model achieves competitive and more balanced performance compared with SOTA methods on the Argoverse motion forecasting benchmark. Comprehensive analyses also indicate that PiH achieves remarkable robustness with a slight 0.2 $\sim$ 0.3 times increase.  As for future work, we plan to extend this idea to the joint prediction and planning field by relying on these faithful interpretable gains.}


\bibliographystyle{named}
\bibliography{ijcai22}

\end{document}